\documentclass[sigconf, nonacm]{acmart}
\usepackage{amsthm}
\usepackage{multirow}
\usepackage{algorithmic}
\newtheoremstyle{user}
  {2pt}   
  {2pt}   
  {\normalfont}
  {0pt}
  {\bfseries}
  {}
  {0pt}
  {\thmnumber{(#2) }}
\theoremstyle{user}

 \newcommand\vldbyear{2026}
\newcommand\vldbworkshop{ADS 2026: The Joint Workshop on Agentic Data Systems and Data-Centric AI (The 1st ADS \& 3rd DATAI)}
\newcommand\vldbauthors{\authors}
\newcommand\vldbtitle{\shorttitle} 
\newcommand\vldbavailabilityurl{https://github.com/BauplanLabs/skill-issues}
\newcommand\vldbpagestyle{empty} 

\usepackage{array}
\usepackage{listings}
\floatstyle{ruled}
\newfloat{listing}{tb}{lst}{}
\floatname{listing}{Listing}

\definecolor{mediumpurple}{rgb}{0.58, 0.44, 0.86}
\definecolor{mediumpurple}{rgb}{0.58, 0.44, 0.86}
\definecolor{MEDIUMPURPLE}{rgb}{0.58, 0.44, 0.86}

\usepackage{tikz}
\usetikzlibrary{backgrounds}
\usetikzlibrary{arrows.meta,positioning}

\tikzset{
  joinop/.style={draw, rounded corners, align=center, font=\scriptsize, inner sep=2.6pt, line width=0.7pt},
  leaf/.style={draw, rounded corners, align=center, font=\scriptsize, inner sep=2.4pt, line width=0.6pt},
  build/.style={-Latex, very thick, black},
  probe/.style={-Latex, thick, dashed, gray!65},
  changed/.style={fill=gray!15, draw=black, line width=0.9pt},
}

\lstdefinelanguage{json}{
    basicstyle=\ttfamily\small,
    showstringspaces=false,
    breaklines=true,
    literate=
     *{:}{{{\color{red}:}}}{1}
      {,}{{{\color{red},}}}{1}
      {\{}{{{\color{blue}\{}}}{1}
      {\}}{{{\color{blue}\}}}}{1}
      {[}{{{\color{blue}[}}}{1}
      {]}{{{\color{blue}]}}}{1}
}

\usepackage[most]{tcolorbox}
\tcbuselibrary{skins,breakable,listings}

\usepackage{siunitx}
\tcbuselibrary{listings,breakable,skins}

\newtcblisting{promptblock}[2][]{%
  breakable,
  enhanced,
  colback=black!2,
  colframe=black!35,
  arc=2pt,
  boxrule=0.5pt,
  left=6pt,right=6pt,top=4pt,bottom=4pt,
  title={#2},
  fonttitle=\bfseries,
  listing only,
  listing options={
    language={},
    basicstyle=\ttfamily\footnotesize,
    columns=fullflexible,
    keepspaces=true,
    breaklines=true,
    showstringspaces=false,
    literate={→}{{$\rightarrow$}}1 {⨝}{{$\bowtie$}}1
  },
  #1
}

\newtcblisting{jsonblock}[2][]{%
  breakable,
  enhanced,
  colback=black!2,
  colframe=black!35,
  arc=2pt,
  boxrule=0.5pt,
  left=6pt,right=6pt,top=4pt,bottom=4pt,
  title={#2},
  fonttitle=\bfseries,
  listing only,
  listing options={
    language=json,
    basicstyle=\ttfamily\footnotesize,
    columns=fullflexible,
    keepspaces=true,
    breaklines=true,
    breakatwhitespace=true,
    showstringspaces=false,
  },
  #1
}
\lstdefinelanguage{SQL}{
  morekeywords={
    SELECT,FROM,WHERE,GROUP,BY,HAVING,ORDER,AS,JOIN,INNER,LEFT,RIGHT,FULL,OUTER,ON,
    WITH,UNION,ALL,DISTINCT,CASE,WHEN,THEN,ELSE,END,AND,OR,NOT,NULL,IS,IN,EXISTS,
    SUM,AVG,MIN,MAX,COUNT,RANK,DENSE_RANK,ROW_NUMBER,OVER,PARTITION,ROWS,RANGE,
    LAG,LEAD,COALESCE,NULLIF,CAST,DATE,EXTRACT,INTERVAL,LIKE,LIMIT,OFFSET
  },
  sensitive=false,
  morecomment=[l]{--},
  morestring=[b]',
}

\lstdefinestyle{sqlpretty}{
  language=SQL,
  basicstyle=\ttfamily\scriptsize,
  columns=fullflexible,
  keepspaces=true,
  breaklines=true,
  breakatwhitespace=true,
  showstringspaces=false,
  tabsize=2,
  upquote=true,
  numbers=left,
  numberstyle=\ttfamily\tiny,
  numbersep=6pt,
  xleftmargin=1.2em,
  keywordstyle=\bfseries\color{black!75},
  stringstyle=\color{black!60},
  commentstyle=\itshape\color{black!45},
}

\definecolor{tpchaccent}{HTML}{CC887A}
\definecolor{tpcdsaccent}{HTML}{7A88CC}

\colorlet{tpchmix}{mediumpurple!40!tpchaccent}    
\colorlet{tpcdsmix}{mediumpurple!40!tpcdsaccent}  

\colorlet{tpchbg}{tpchmix!10!white}
\colorlet{tpchframe}{tpchmix!60!black}

\colorlet{tpcdsbg}{tpcdsmix!10!white}
\colorlet{tpcdsframe}{tpcdsmix!60!black}

\newtcblisting{tpchquery}[2][]{%
  breakable,
  enhanced,
  colback=tpchbg,
  colframe=tpchframe,
  arc=2pt,
  boxrule=0.5pt,
  left=6pt,right=6pt,top=4pt,bottom=4pt,
  title={#2},
  fonttitle=\bfseries\small,
  listing only,
  listing options={
    style=sqlpretty
  },
  #1
}

\newtcblisting{tpcdsquery}[2][]{%
  breakable,
  enhanced,
  colback=tpcdsbg,
  colframe=tpcdsframe,
  arc=2pt,
  boxrule=0.5pt,
  left=6pt,right=6pt,top=4pt,bottom=4pt,
  title={#2},
  fonttitle=\bfseries\small,
  listing only,
  listing options={
    style=sqlpretty
  },
  #1
}

\usepackage{subcaption}
\usepackage{xcolor} 
\usepackage[normalem]{ulem}
\usepackage{float}

\newlength{\algcommentwidth}
\newlength{\algcodewidth}
\newlength{\algcommentprobe}  
\newlength{\algcommentcol}    

\renewcommand{\algorithmiccomment}[1]{%
  \setlength{\algcommentcol}{0.45\linewidth}%
  \settowidth{\algcommentprobe}{\footnotesize$\triangleright$~#1\ }%
  \ifdim\algcommentprobe>\algcommentcol
    \unskip\hfill\penalty-10000
    \hspace*{3.5em}%
    \begin{minipage}[t]{\dimexpr\linewidth-4em\relax}%
      \footnotesize\raggedright
      \setlength{\parindent}{0pt}%
      \leftskip=1em \hangindent=-1em \hangafter=1
      \noindent\llap{$\triangleright$~}#1\par
    \end{minipage}%
    \vspace{0.6ex}%
  \else
    \quad\hbox to \dimexpr\algcommentcol-\algcommentprobe\relax{}%
    {\footnotesize$\triangleright$~#1}%
  \fi
}

\definecolor{suggestcolor}{rgb}{0.10,0.45,0.85}

\newif\ifshowsuggestions
\showsuggestionsfalse

\ifshowsuggestions
  
\else
  
\fi
\definecolor{deletecolor}{rgb}{0.85,0.20,0.20}
\ifshowsuggestions
  \newcommand{\delete}[1]{\textcolor{deletecolor}{\sout{#1}}} 
\else
  \newcommand{\delete}[1]{} 
\fi

\providecommand{\citet}[1]{\citeauthor{#1}~\cite{#1}}

\newtcolorbox{schemabox}[2][]{
  lower separated=false,
  colback=white,
  colframe=mediumpurple,
  fonttitle=\bfseries\small,
  colbacktitle=mediumpurple,
  coltitle=white,
  enhanced,
  attach boxed title to top left={yshift=-0.07in,xshift=0.1in},
  boxed title style={boxrule=0pt,colframe=white},
  title=#2,#1}

\newtcolorbox{tracebox}[2][]{breakable,enhanced,colback=black!2,colframe=black!35,arc=2pt,boxrule=0.5pt,left=6pt,right=6pt,top=4pt,bottom=4pt,title={#2},fonttitle=\bfseries,#1}

\newbool{showcomments}
  \boolfalse{showcomments}

\begin{document}
\title{``Skill Issues'': Data-Centric Optimization of Lakehouse Agents}
\author{Nicole Rose Schneider}
\email{nsch@umd.edu}
\affiliation{%
  \institution{University of Maryland}
  \country{USA}
}

\author{Davide Ghilardi}
\email{davide.ghilardi@unimib.it}
\affiliation{%
  \institution{Università Milano Bicocca}
    \country{Italy}
}

\author{Giacomo Piccinini}
\email{giacomo.piccinini@bauplanlabs.com}
\affiliation{%
  \institution{Bauplan Labs}
    \country{USA}
}

\author{Jacopo Tagliabue}
\email{jacopo.tagliabue@bauplanlabs.com}
\affiliation{%
  \institution{Bauplan Labs}
  \country{USA}
}

\begin{abstract}
Coding agents are becoming users of data infrastructure, but their success depends not only on model quality: it also depends on the skills and environment files that teach agents how to use a system. We study how to optimize these artifacts for agents operating on a branching lakehouse, \texttt{Bauplan}. In our setting, headless APIs and Git-like data primitives expose data workflows through code, branches, commits, and merges. Our central observation is that a branching lakehouse turns data-agent evaluation from an output-matching problem into a state-verification problem: agent-generated pipeline code induces concrete, inspectable lakehouse changes. We present a data-centric optimization pipeline that generates task-verifier pairs, executes candidate skills in isolated sandboxes, and scores trajectories using both trace-level signals and programmatic checks over lakehouse state. In a preliminary evaluation on hundreds of tasks, optimized skills improve held-out reward by up to 28.6\%. These results suggest that write-path data workflows provide a useful substrate for optimizing agent skills beyond read-only tasks.
\end{abstract}

\maketitle

\pagestyle{\vldbpagestyle}
\begingroup\small\noindent\raggedright\textbf{VLDB Workshop Reference Format:}\\
\vldbauthors. \vldbtitle. VLDB \vldbyear\ Workshop: \vldbworkshop.\\ 
\endgroup
\begingroup
\renewcommand\thefootnote{}\footnote{\noindent
This work is licensed under the Creative Commons BY-NC-ND 4.0 International License. Visit \url{https://creativecommons.org/licenses/by-nc-nd/4.0/} to view a copy of this license. For any use beyond those covered by this license, obtain permission by emailing \href{mailto:info@vldb.org}{info@vldb.org}. Copyright is held by the owner/author(s). Publication rights licensed to the VLDB Endowment. \\
\raggedright Proceedings of the VLDB Endowment. 
ISSN 2150-8097. \\
}\addtocounter{footnote}{-1}\endgroup

\ifdefempty{\vldbavailabilityurl}{}{
\vspace{.3cm}
\begingroup\small\noindent\raggedright\textbf{VLDB Workshop Artifact Availability:}\\
The source code, data, and/or other artifacts have been made available at \url{https://github.com/BauplanLabs/skill-issues}.
\endgroup
}

\section{Introduction}

\begin{quotation}
``Make something idiot proof, and somebody will come up with a better idiot.'' \textit{Anonymous}
\end{quotation}

Whether they work synchronously with humans \cite{huang_control_2025} or asynchronously in a loop \cite{yao2023reactsynergizingreasoningacting}, it is clear that coding agents will soon become the primary users of cloud infrastructure. Experts point out that the affordances in traditional Online Analytical Processing (\textit{OLAP}) systems make agents ineffective and unsafe on the lakehouse; \texttt{Bauplan} was designed as an agent-first data platform according to the ``AI Overlord Manifesto'' \cite{liu2025supportingaioverlordsredesigning}: fast, safe execution coupled with Git-like collaboration primitives \cite{sheng2026buildingcorrectbydesignlakehousedata,sheng2026gitlakegitfordataagenticlakehouse}.

Running \texttt{Bauplan} gives us a privileged perspective on the challenges of scaling data agents to production settings. Even as the raw performance of Large Language Models (\textit{LLM}s) increases, the variance in agentic success is huge: \texttt{Bauplan} APIs are not yet well represented in the training set, users' prompting abilities vary widely and the new affordances call for revised best practices. Following common LLM-agent practice \cite{ant_skill}, we shared standardized Markdown files (\textit{skills}) with users to both lower the barrier to using the platform \textit{and} ensure a consistent standard in how code is written, tested, and deployed.

Customer choices in terms of models, coding assistants, and usage patterns have evolved quickly in recent months, so manually curating skills is no longer a sustainable process. More importantly, we found that this is not merely a prompt-engineering problem: when agents operate on production data systems, the textual artifacts surrounding them become part of the system control surface, and need to be optimized and evaluated as such. As it turns out, turning our initial prompting strategy into a scalable and reliable process required assembling a complex and multi-faceted set of capabilities: data mining over traces, user interviews on domain knowledge, LLM-as-judge calibration and more. In \textit{this} paper, we share our skill optimization journey with the community. In particular, we summarize our contributions as follows:

\begin{enumerate}
    \item through the privileged lens provided by running thousands of agentic workloads per day on \texttt{Bauplan}, we motivate the importance of skills to support common multi-turn agentic use cases seen in production, such as ETL, Write-Audit-Publish, and data exploration;
    \item we share the optimization loop for \texttt{Bauplan}'s skills, with gains of up to 28.6\% relative to the seed skills. Skills associated with complex, multi-skill workflows improved only marginally when optimized in isolation, motivating future work on joint skill optimization. Since closed-source LLMs are frozen, and optimization frameworks are now common, we embrace the data-centric AI methodology \cite{zha2023data} and focus on the surrounding data tasks. Great care is taken in the dataset generation process, as a result of trade-offs among the nuances of synthetic data generation, production traces, data engineering knowledge and error analysis. We release the full setup to the community, stripped of \texttt{Bauplan}-specific features, as an open-source contribution;
    \item we highlight the interplay between \texttt{Bauplan} affordances, the optimization loop and the final skills. For example, a statically checkable SDK reduces round-trips; progressive CLI discovery allows for concise skills; more importantly, the isomorphism between pipeline \textit{code} and \textit{data commits} grounds fine-grained verification of write-path behavior (Section~\ref{sec:data-generation-verification}).
\end{enumerate}

Since unsupervised workloads on production data are dangerous in most OLAP systems \cite{liu2025supportingaioverlordsredesigning}, lakehouse agents today are still in their infancy and mostly focused on individual tasks in the \textit{read} path (text-to-SQL \cite{liu2025surveytexttosqlerallms}, data exploration \cite{chen2025largelanguagemodelbaseddata}). While our (reproducible) experimental setup is tailored to a specific set of APIs, our strongest contribution is linking Git-for-data to the optimization of compound AI systems for OLAP use cases: as such, alternative implementations of the same ideas will benefit from the insights we share. For these reasons, we believe our work to be useful to a wide set of practitioners: both as a reference implementation of a full optimization pipeline and as an example of a realistic coding setup specifically targeted at the full life-cycle of data management, including potentially disruptive operations such as \textit{writes} and \textit{deletes}. In contrast to evaluations that stop at read-only SQL answers, our setting requires checking whether an agent changed the lakehouse state in the intended way.
\section{Setup and problem statement}

\subsection{The agentic setup}

\begin{figure}
    \centering
    \includegraphics[width=\columnwidth]{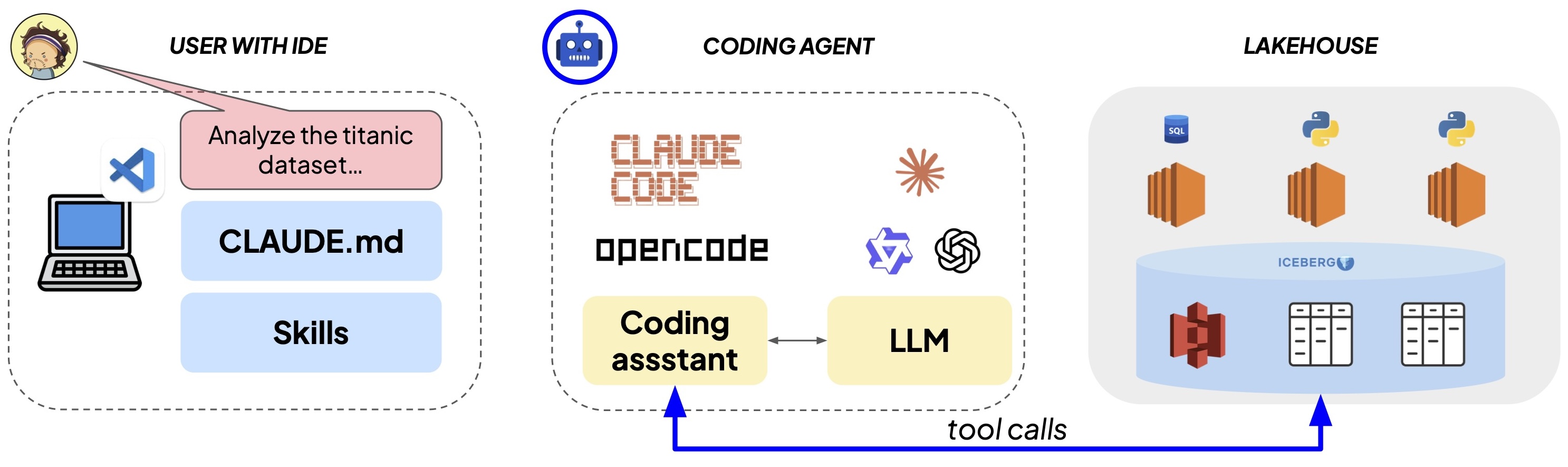}
    \caption{Coding agent setup. The user chooses a task and provisions a coding agent (e.g. Claude Code), which in turn is wired to a Large Language Model (e.g. Opus): tool calls (Bash or Python execution) invoke lakehouse-specific APIs and feed the result back to the agent. \textit{CLAUDE.md} and \textit{Skills} are the only ``hyperparameters'' under our control: how do we optimize them?}
    \label{fig:setup}
\end{figure}

Fig.~\ref{fig:setup} illustrates the standard setup for users of \texttt{Bauplan}. When faced with a data task, such as analyzing data, building pipelines end-to-end, or ingesting data safely, a user can pick a coding assistant (e.g. \textit{Claude Code} \cite{liu2026diveclaudecodedesign}) wired to \textit{Opus} and ask it to perform the task. Following industry best practices \cite{liang2026skillnetcreateevaluateconnect,ant_skill}, we provide the user with \textit{skills} and a \textit{CLAUDE.md}\footnote{In a non-Anthropic setup, we use the cross-provider \textit{AGENTS.md} standard instead: for simplicity, we use the Claude-specific lexicon throughout the paper.} to improve performance on data-specific scenarios. Skills prepare the agent to solve a problem ``by injecting procedural knowledge, modifying execution context, and enabling progressive disclosure of information'' \cite{xu2026agentskillslargelanguage}. Note that the agent could also work fully asynchronously: the same user choices and degrees of freedom apply, but instructions are possibly even more important there because of the autonomy in execution.

When optimizing this setup against a series of tasks, ``backpropagation'' can only happen at the level of these textual hyperparameters, as everything else is held constant by user choices. Our research question is then easily stated: how can we optimize these parameters -- \textit{skills} and \textit{CLAUDE.md} -- to improve the performance of the ``compound AI system'' \cite{chen2025optimizingmodelselectioncompound} on data use cases?

Answering this question will involve generating a dataset of use cases (Section~\ref{sec:data-generation}), scoring outcomes at scale (Section~\ref{sec:data-generation-verification}), and choosing an optimization method (Section~\ref{sec:optimization}). Before diving into these details, we provide a small introduction to \texttt{Bauplan}, as the platform ergonomics not only exemplify its agentic nature, but also significantly shape the optimization process itself.

\subsection{Bauplan overview}
\label{sec:bauplan}

\texttt{Bauplan} is an agent-first lakehouse, built on the separation of storage (Apache Iceberg \cite{iceberg} on S3) and compute (SQL and Python). We refer the reader to \cite{10.1145/3702634.3702955} for a full description, and survey here two distinctive features of \texttt{Bauplan} (vis-a-vis a traditional lakehouse), which are particularly relevant for a successful verification process (Section~\ref{sec:data-generation-verification}):

\begin{enumerate}
    \item \textbf{headless}: \textit{all} interactions with the platform (running jobs, creating tables, inspecting logs, etc.) are API-based, as wrapped by a CLI (for Bash use cases) and a Python SDK (for scripting use cases); this matters because it allows agents to manage the \textit{full data life-cycle} through code, and because it allows us to programmatically verify lakehouse changes at scale;
    \item \textbf{git-for-data}: \texttt{Bauplan} re-implements Git primitives over data assets, allowing \textit{commits}, \textit{branches}, and \textit{merges} on lakehouse tables \cite{10.1145/3650203.3663335}. Git-like APIs allow safe, concurrent changes and prevent unverified AI code from reaching production; this matters because it allows us to deterministically map API calls to data changes, which in turn is crucial for verification.
\end{enumerate}

To get a sense of the code-first nature of the platform and to understand the \textit{effects} of a run, we can inspect two snippets, a DAG and a run orchestration. The first is a one-node DAG taking \texttt{source} (an Iceberg table) as input, and producing \texttt{table-1} through function \texttt{f}:

\begin{lstlisting}[
  language=Python,
  showstringspaces=false,
  columns=fullflexible,
  caption={A one-node Python DAG.},
  label={lst:dag},
  basicstyle=\ttfamily\scriptsize,
  numbers=none
]
class Total(BauplanSchema):
    category: str
    total_amount: float

@model(name="table-1", materialize="REPLACE")
@python("3.12", pip={"polars": "1.27"})
def f(df=source) -> Total:
    tbl = df.group_by("category").agg(pl.col("amount").sum())
    return tbl
\end{lstlisting}

The abstractions are hopefully self-ex\-plan\-a\-tory. We only briefly highlight the declarative nature of both infrastructure and I/O, which is an ideal target for code generation. \textit{Physical} operations (installing packages, reading data from S3, or writing data to S3) happen in ``platform space'', so that user space only expresses \textit{logical} operations and transformation code is executed in secure and sandboxed compute \cite{10825377}. Importantly, DAGs live in a complex lakehouse environment, which is \textit{also fully programmable}. Interactive use cases are typically managed from the CLI (e.g. \texttt{bauplan branch create feature-1}, \texttt{bauplan run}), but complex flows are built by interleaving lakehouse commands and Python control flow, leveraging a fully typed SDK \cite{montana2026notyourusualtypes}:

\begin{lstlisting}[
  language=Python,
  showstringspaces=false,
  columns=fullflexible,
  caption={Python SDK example.},
  label={lst:api},
  basicstyle=\ttfamily\scriptsize,
  numbers=none
]
# 1) instantiate the client
client = bauplan.Client()
# 2) create a development branch from main
dev_br: Branch = client.create_branch("dev_br", from_ref="main")
# 3) run a DAG on the branch
run_state = client.run("pipeline/", ref=dev_br)
# 4) merge the branch into production (on success)
if run_state.success():
    client.merge(dev_br, into="main")
    client.delete_branch(dev_br)
\end{lstlisting}

Listing~\ref{lst:api} is the ``outer loop'' for Listing~\ref{lst:dag}, executing the transformation defined there: again, the abstractions should be self-ex\-plan\-a\-tory, and we only highlight that runs are embedded in a data branching workflow that is very similar to the typical Git flow for code.

\begin{figure}
    \centering
    \includegraphics[width=\columnwidth]{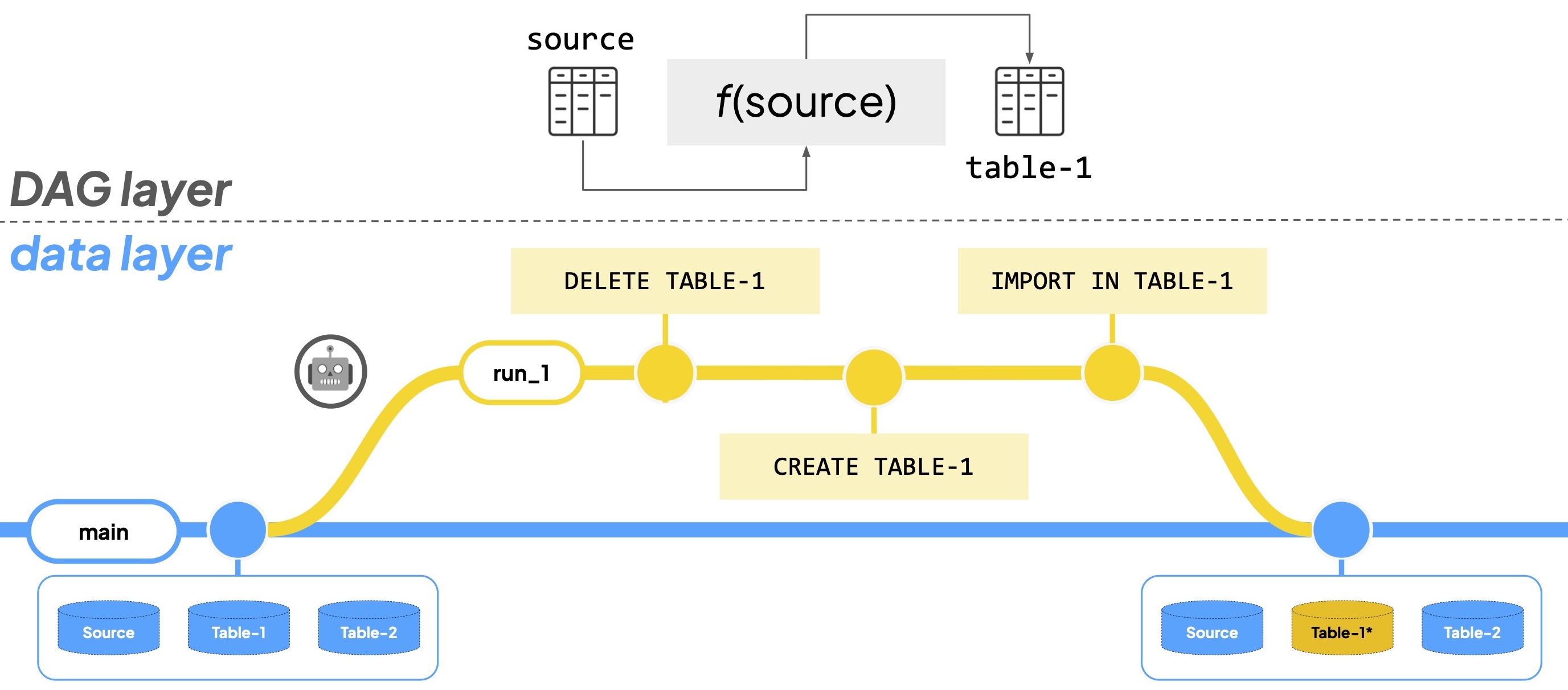}
    \caption{Pipeline code maps to \textit{verifiable} lakehouse changes. A one-node DAG transforming \textit{source} into \textit{table-1} (Listing~\ref{lst:dag}) runs on a data branch: a successful run will alter the lake state in a predictable and verifiable way. Here, commits in the branch correspond to \textit{writes} (delete, create, import), and one commit in \texttt{main} corresponds to the \textit{merge}, which atomically updates only the snapshot of \textit{table-1}. The same APIs can be used by a verifier to check data changes commit by commit.}
    \label{fig:iso}
\end{figure}

What happens when \texttt{client.run} above gets called? The code is shipped to cloud workers, and logs are streamed back for the agent to observe and react \cite{10.1145/3702634.3702955}. Crucially, every \textit{write} in the execution corresponds to a data \textit{commit}, i.e. an immutable, Git-like reference to the state of the lakehouse at that time \cite{sheng2026buildingcorrectbydesignlakehousedata}. Fig.~\ref{fig:iso} illustrates the code-data isomorphism: any actor (human or LLM) reading Listing~\ref{lst:dag} should be able to enumerate what a successful run would entail in the shape of \textit{commits} that mark how the lake changes. When we ask an LLM to generate task-verifier pairs (Section~\ref{sec:data-generation}), this allows us to \textit{precisely} verify whether the agent did indeed perform the work as expected, minimizing ``false positives'' and cheating behavior (Section~\ref{sec:data-generation-verification}). This is the core difference between optimizing skills for a branching lakehouse and generic prompt engineering: the system exposes concrete write-path effects that can be checked after execution.

\section{Skill optimization}

\begin{table*}[ht]
  \centering
\caption{Lifecycle of one benchmark task in the optimization harness.}
\label{tab:agent-run-steps}
  \begin{tabular}{c l l}
  \hline
  \# & Step & What it does \\
  \hline
  1 & Create \texttt{<username>.main} & Branches the agent's writable \texttt{<username>.main} off \texttt{main}. \\
  2 & Prep initial state & Drops any tables that are explicitly required by the dataset entry to be absent on \texttt{<username>.main}. \\
  3 & Capture baseline & Snapshots the latest commit on \texttt{main} and \texttt{<username>.main}. \\
  4 & Agent runs & Runs the Claude agent on the task prompt. \\
  5 & Validation & Scores the result against the dataset entry's \texttt{end\_conditions}. \\
  6 & Teardown & Deletes post-baseline \texttt{<username>.*} branches, then drops \texttt{<username>.main}. \\
  \hline
  \end{tabular}
  \end{table*}

\subsection{Architecture}

\begin{figure}
    \centering
    \includegraphics[width=\columnwidth]{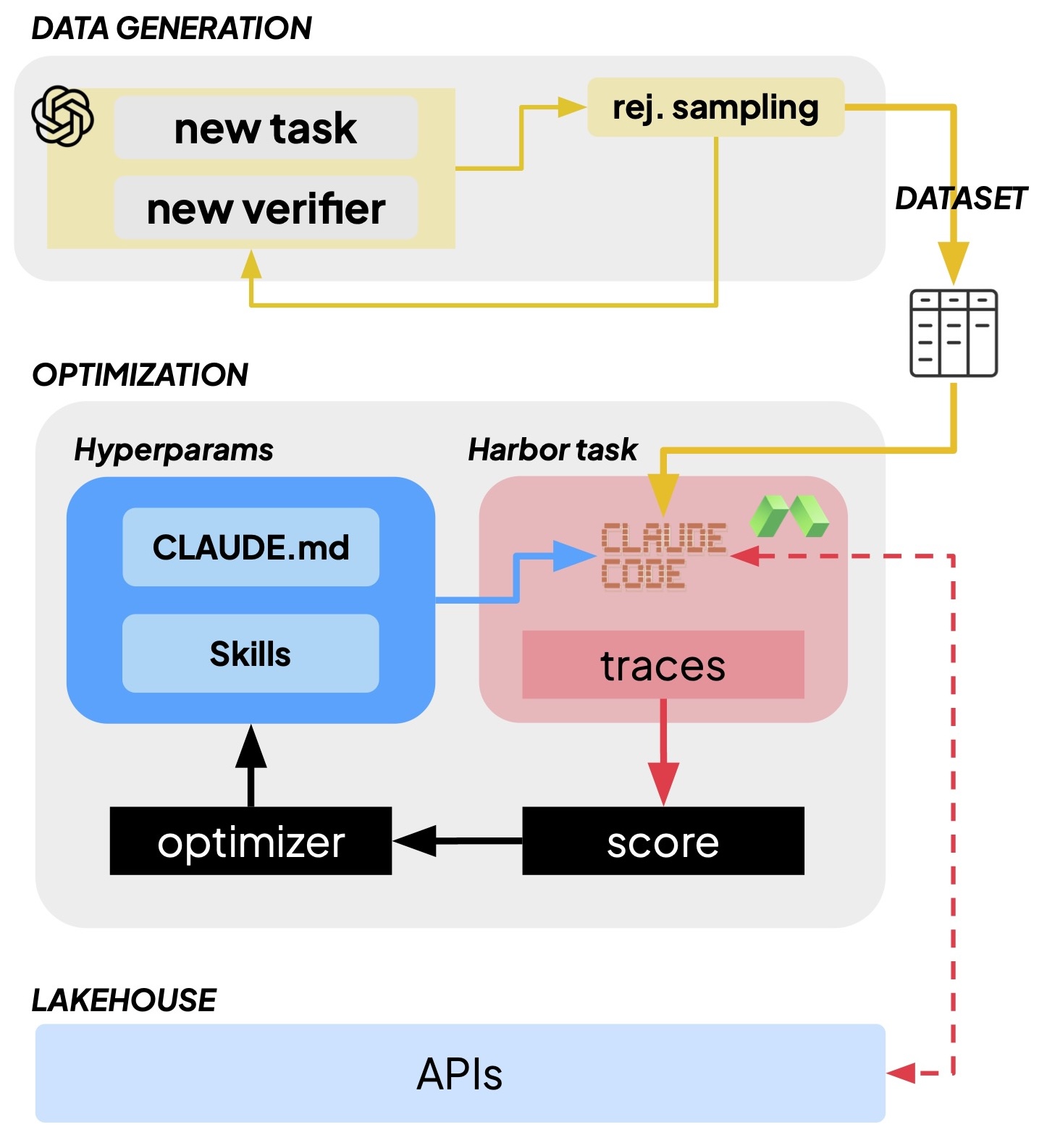}
    \caption{High-level system architecture. An LLM-powered data generation module builds a dataset with \textit{task-verifier} pairs. The optimization loop runs the target coding setup on the tasks with the current \textit{skills} -- tools connect to \texttt{Bauplan} to execute actions and get feedback. Metrics are then computed and new skills are produced by the optimizer, and the loop continues.}
    \label{fig:arch}
\end{figure}

Fig.~\ref{fig:arch} shows the high-level architecture of the optimization pipeline. First, an LLM-powered data generation module (Section~\ref{sec:scenario-generation}) takes as input a taxonomy of scenarios (data ingestion, report building, pipeline debugging etc.) and personas (data engineer, data scientist, data analyst etc.) and \texttt{Bauplan}-specific prompt instructions to generate \textit{task-verifier} pairs (Section~\ref{sec:data-generation-verification}): as discussed (Fig.~\ref{fig:iso}), a DAG that would solve the task can be mapped deterministically to lakehouse checks, so that tasks can have fine-grained, programmatic verification.

Once the dataset is built, we use Harbor~\cite{Harbor_Framework} to orchestrate Modal sandboxes initialized with Markdown files and the appropriate coding assistant package. Table~\ref{tab:agent-run-steps} summarizes the life-cycle of each
benchmark task: the harness creates an isolated writable branch, captures the baseline lakehouse state, lets the agent act, validates the outcome, and tears down any temporary state. This fixed life-cycle is what allows the same optimization loop to evaluate both read-only and write-path tasks safely. Traces collected from these trajectories, as well as a final score, are then fed into an optimization function, which generates new Markdown files for another round.

\subsection{Data generation}
\label{sec:data-generation}

The optimization loop in Fig.~3 requires a benchmark that is both realistic enough to exercise the skills used in production, and structured enough to support automatic verification. We therefore generate \emph{task-verifier pairs}: each benchmark entry contains not only a user-facing prompt, but also the initial lakehouse state, the expected platform interactions, forbidden actions, response-level checks, and a programmatic validation script. This follows a data-centric view of agentic improvement~\cite{zha2023data}, with a strong focus on quality \textit{at scale}: while human judgment and domain knowledge play a role in the initial scenario design, the final dataset is generated in a loop until it satisfies the structural and semantic constraints required by the downstream optimization task.

\subsubsection{Scenario generation}
\label{sec:scenario-generation}

Table~\ref{tab:api-abstractions} reports representative counts for compute and branch operations as collected from a May 2026 sample of anonymized production logs. The table provides an empirical prior over agentic workflows: imports, queries, and pipeline runs all occur frequently on the compute side, while branch deletion and merge operations dominate the branch-management side. This confirms that realistic agent workloads are not limited to read-only SQL, but include useful but potentially dangerous write-path operations. Crucially, we note that raw frequencies are not a sufficient benchmark specification. First, traces are partially shaped by the current skills, so they may over-represent APIs described well and under-represent poorly documented behaviors. Second, some rare operations are semantically important for safety. We therefore use the counts as a ``conceptual prior'' for scenario design, not as the final task mixture: common APIs should appear often enough to reflect real usage, but rare APIs must still be covered by construction. 

Notably, looking only at API counts flattens multi-step behavior, so other important priors for scenario design are \textit{sessions}, i.e. series of commands issued when performing a coherent multi-step process. In particular, improvements in LLMs' theory of mind \cite{chen2025theorymindlargelanguage} set the stage to perform inverse planning~\cite{baker2009action,gelp2025machinetheorymindlarge} from session-level API traces; e.g., a sequence of branch creation, import, validation query, and merge suggests a safe ingestion workflow. A model fed with sessions can infer the kind of intention a user may have had, and these inferred intentions become templates for task generation. Putting together API counts and session-level insights, the final dataset is synthetic in form, but anchored in real platform usage and in the operational semantics of the lakehouse.

The resulting taxonomy has two axes. The first is the type of lakehouse work: read-only tasks, such as data exploration, text-to-SQL, and data-management inspection; and write-path tasks, such as safe ingestion, new pipeline construction, adding expectations, modifying or debugging pipelines, branch operations, and end-to-end analysis. The second axis is the shape of the user request. Each task is tagged with a difficulty level and a prompt-specificity level: vague prompts model novice users, medium prompts model typical practitioners, and detailed prompts model expert users who specify tables, columns, output schemas, and acceptance criteria. This is important because the skills must help agents not only when the desired workflow is explicit, but also when the user request has to be translated into the right workflow. This approach may be seen as a further natural generalization of taxonomy-based dataset generation \cite{davidson2026reasoningdrivensyntheticdatageneration}: on top of generating tasks by progressively refining intent into scenarios and then into tasks (``depth''), and by progressively covering the design space with additional variations (``coverage''), we also generate tasks by varying a user-specific dimension (``explicitness'').

Given this taxonomy, an LLM generator emits one structured benchmark entry at a time through a typed tool call. The generator is grounded with Bauplan-specific context (skills, SDK, CLI) and the available datasets: for ingestion tasks, the prompt refers to the S3 URI directly; in any other case, the entry declares any pre-loaded tables. Each entry is syntactically validated and then passed through a thorough set of checks: e.g. commands must correspond to real CLI arguments. For read-only categories, the validator requires at least one write-blocking forbidden action and at least one no-mutation state assertion, preventing read tasks from silently allowing destructive behavior. Invalid entries are rejected and the generator samples again until enough valid tasks are collected~\cite{wang2023selfinstruct}. Before optimization begins, domain experts carefully sample tasks and evaluate their quality in detail, to verify both prompt realism (i.e., is this a realistic user scenario and a properly worded request?) and verifier thoroughness (i.e. do expected commands and SDK checks prove that the agent executed what the user wanted?). We refer readers to the open-source repository and to Appendix~\ref{app:examples} for full-fledged examples. 

\subsubsection{Verification}
\label{sec:data-generation-verification}

Our harness verifies each trajectory at three levels: tool traces, lakehouse state, and final response quality. This is conceptually similar to executable software-engineering benchmarks such as SWE-bench~\cite{jimenez2024swebench}, where candidate code changes are evaluated by running tests. In our setting, however, the executable object is not only a source-code patch: it is a sequence of agent actions whose effects include branches, commits, tables, schemas, rows, and merges. We refer the reader to the repository for the full list of checks, and survey here the high-level features of each category: 

\begin{itemize}
    \item \textbf{Trace-level verification}: the verifier checks whether the agent used the expected skills and tool calls, and avoided forbidden actions (e.g. \texttt{bauplan run} in read-only). Expected operations are checked against both Bash calls and Python scripts (with the appropriate CLI-to-SDK syntax conversion): in most cases, the model is left free to choose between the CLI and the SDK with no penalties.
    \item \textbf{Lakehouse-state verification}: the verifier checks the \textit{data-specific} preconditions and postconditions. For example, simple assertions cover branch existence and no-change guarantees, while more complex programmatic checks validate runs commit-by-commit (Fig.~\ref{fig:iso}). For write-path tasks whose final state depends on the user name, the validation script uses the SDK to discover the username dynamically and assert the intended state. Based on our experience, only frontier closed models such as GPT-5.5 were able to properly ``one-shot'' the entire task generation with the required internal consistency (at the cost of high latency): we leave the exploration of smaller models and multi-agent orchestration to future work.
    \item \textbf{Response-level verification}: some tasks do not modify the lakehouse state, nor do they have a structured output: exploration, for example, typically ends with a free-form English response. The verifier therefore combines hard and soft checks. Hard constraints are deterministic checks, e.g. string matching and regex. Soft constraints are rubric-style criteria graded by an LLM judge~\cite{zheng2023judging}, checking whether the answer covers the requested dimensions and states the right caveats.
\end{itemize}

While this level of detail-oriented verification and thoroughness may seem unnecessary for special-purpose data agents, in our view this approach is the heart of the optimization process. Since LLMs and optimization libraries can be (and often are) black-box components, iterating on data quality provides the needed leverage to steer model behavior on real-world, niche scenarios. Two Bauplan features (Section~\ref{sec:bauplan}) proved to be crucial for the data-centric methodology: first, the SDK allows verifiers to be written using the same methods that a data agent would use; second, git-for-data makes write-path evaluation deterministic and fine-grained: a candidate skill is rewarded only when the agent changes the lakehouse in the intended way.

\begin{table}[t]
\centering
\small
\caption{Usage of \texttt{Bauplan}'s abstractions in a May 2026 sample of anonymized production logs. Percentages are computed within each API group and rounded to whole numbers.}
\label{tab:api-abstractions}
\begin{tabular}{p{0.26\columnwidth} p{0.46\columnwidth} r}
\toprule
\textbf{Abstraction} & \textbf{Description} & \textbf{Pct.} \\
\midrule
\multicolumn{3}{l}{\textbf{Compute APIs}} \\
Import &
Append data to a table &
40\% \\
Query &
Execute a SQL query &
35\% \\
Run &
Run a pipeline &
25\% \\
\midrule
\multicolumn{3}{l}{\textbf{Branch APIs}} \\
Merge &
Merge a ref into a target branch &
43\% \\
Delete &
Delete a branch &
37\% \\
Create &
Create a new branch from a ref &
11\% \\
Get &
Get details about a specific branch &
8\% \\
List &
List branches &
1\% \\
\bottomrule
\end{tabular}
\end{table}

\subsection{Optimization}
\label{sec:optimization}

We used GEPA Optimize Anything~\cite{Agrawal2025gepa} to optimize the text contents of the skills. Inputs to the optimization function are
\begin{enumerate}
    \item A seed skill
    \item An evaluation function
    \item A description of the goal
\end{enumerate}

The seed skill is the initial skeleton description for the skill that we aim to optimize.
The evaluation function takes candidate skills generated by GEPA and runs a sandboxed agent on dataset tasks that involve the relevant skill.
The validation is done after the agent runs, and it is used to score the pipeline with a reward between 0 and 1, based on how many validation checks pass.
Based on the reward score and any error traces generated as actionable side information, GEPA generates new candidates and the process continues.

Importantly, the architecture is designed to be modular (Fig.~\ref{fig:arch}): as long as modules respect the declared interfaces (see the open source repository), it is easy to plug in different algorithms keeping the rest constant. We leave experimentation with different frameworks~\cite{novikov2025alphaevolvecodingagentscientific,liu2026evoxmetaevolutionautomateddiscovery} to future work.

\begin{table}[h]
  \centering
  \caption{Dataset composition by type and scenario.}
  \label{tab:dataset}
  \setlength{\tabcolsep}{5pt}
  \begin{tabular}{llr}
  \toprule
  Scenario & Task type & Count \\
  \midrule
  \multirow{1}{*}{Read}
  & Exploration            & 139 \\
  & Assessment & 126 \\
  & Data quality check & 100 \\
  \midrule
  \multirow{3}{*}{Write}
  & New pipeline     & 209 \\
  & Fix table       & 100 \\
  & Ingestion            & 106 \\
  \midrule
  \textbf{Total} & & \textbf{780} \\
  \bottomrule
  \end{tabular}
\end{table}


%
%
\begin{figure*}[htbp]
  \centering
  \begin{subfigure}[t]{0.3\textwidth}
    \includegraphics[width=\linewidth]{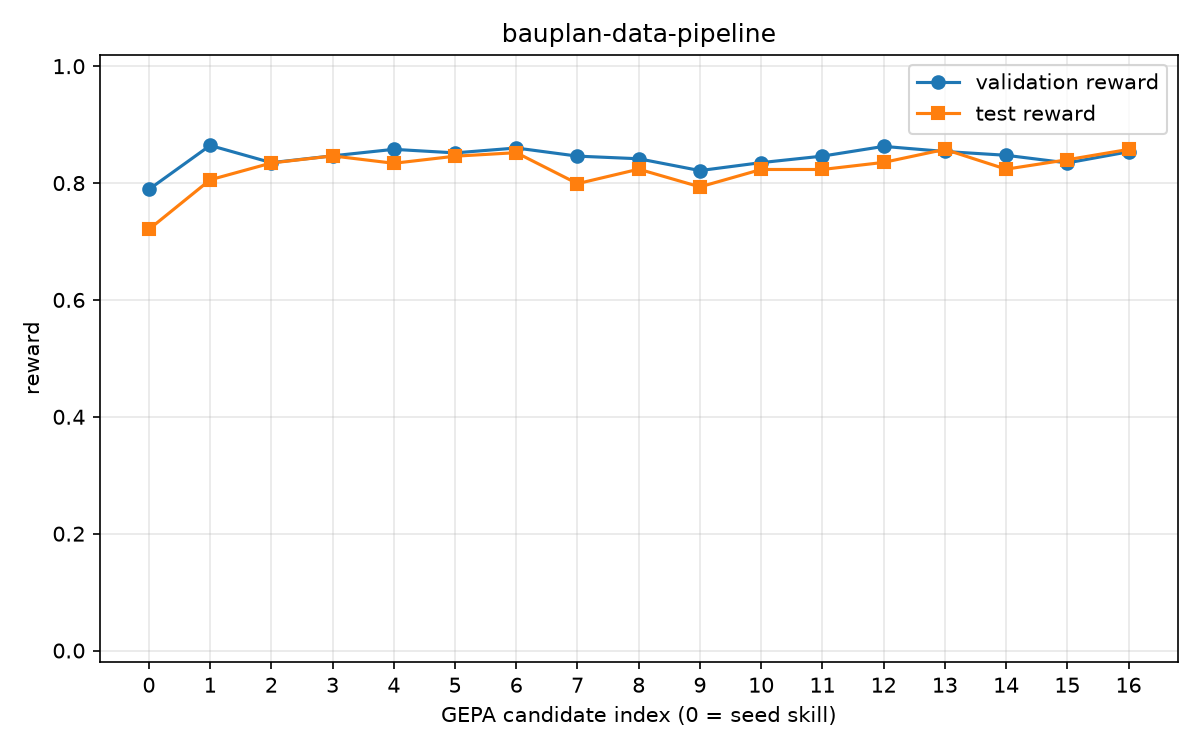}
    \caption{bauplan-data-pipeline}
    \label{fig:skill4-data-pipeline}
  \end{subfigure}%
  \hfill%
  \begin{subfigure}[t]{0.32\textwidth}
    \includegraphics[width=\linewidth]{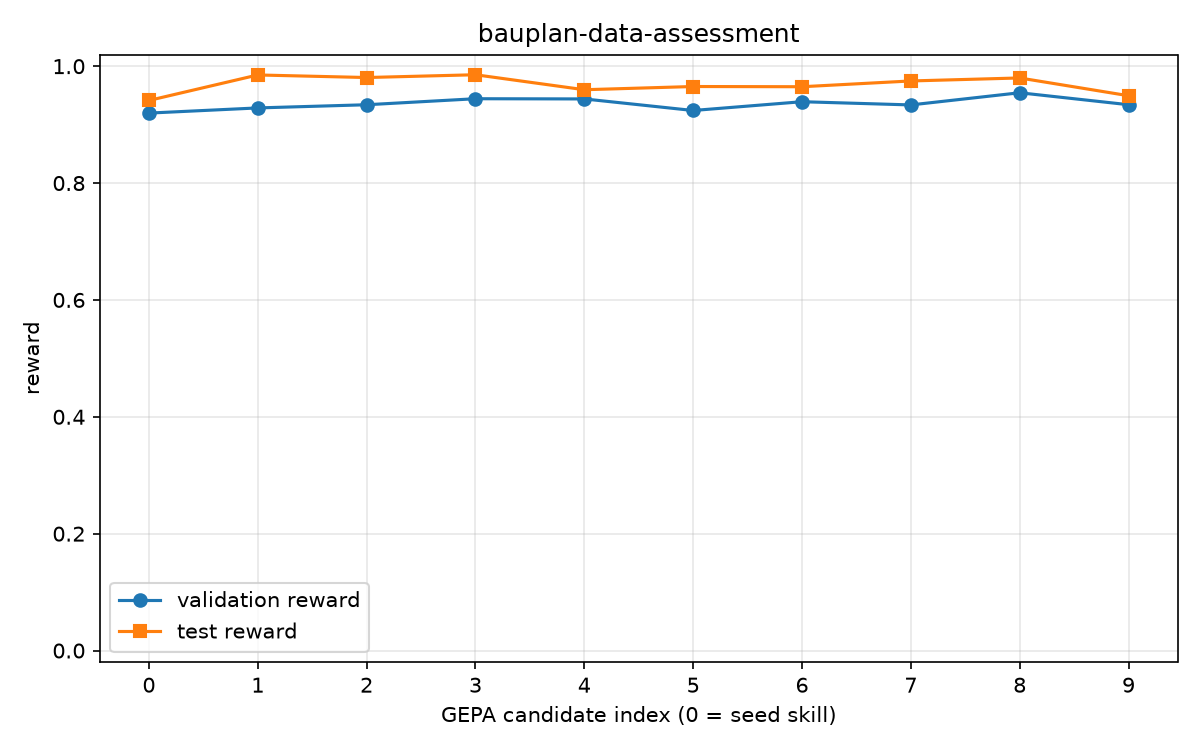}
    \caption{bauplan-data-assessment}
    \label{fig:skill4-data-assessment}
  \end{subfigure}
\hfill%
  \begin{subfigure}[t]{0.32\textwidth}
    \includegraphics[width=\linewidth]{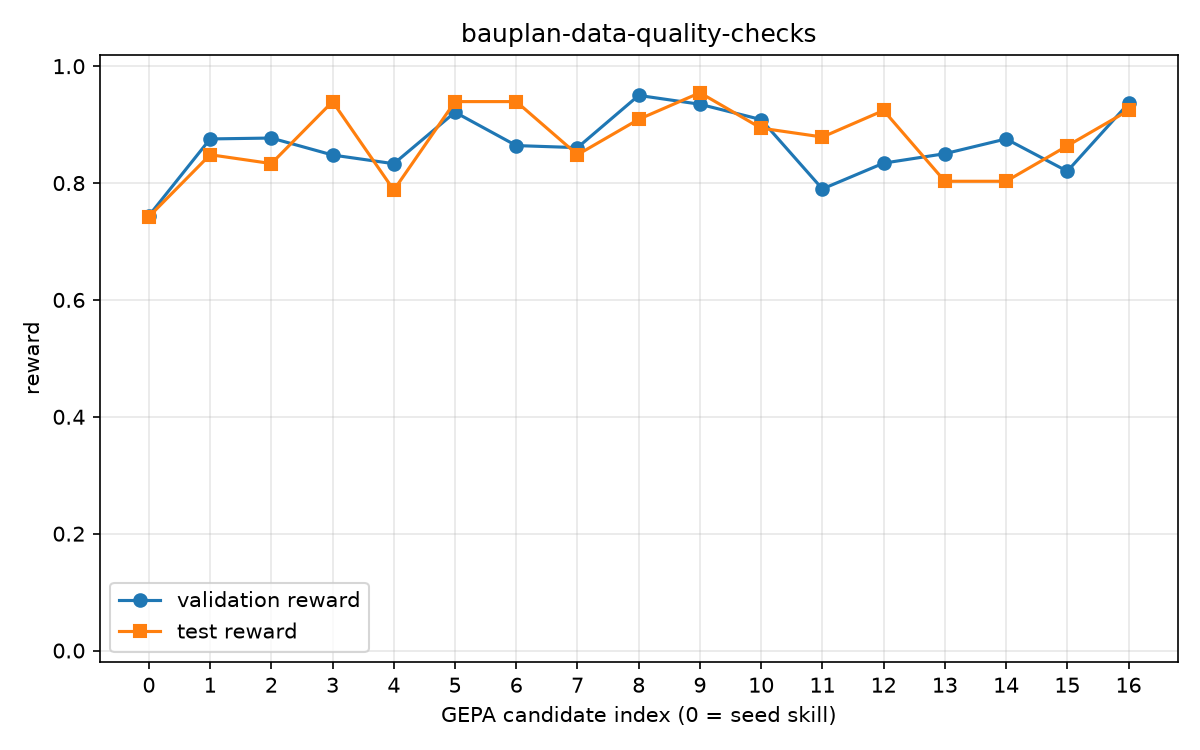}
    \caption{bauplan-data-quality-checks}
    \label{fig:skill4-data-quality-checks}
  \end{subfigure}%
  \hfill
  \begin{subfigure}[t]{0.32\textwidth}
    \includegraphics[width=\linewidth]{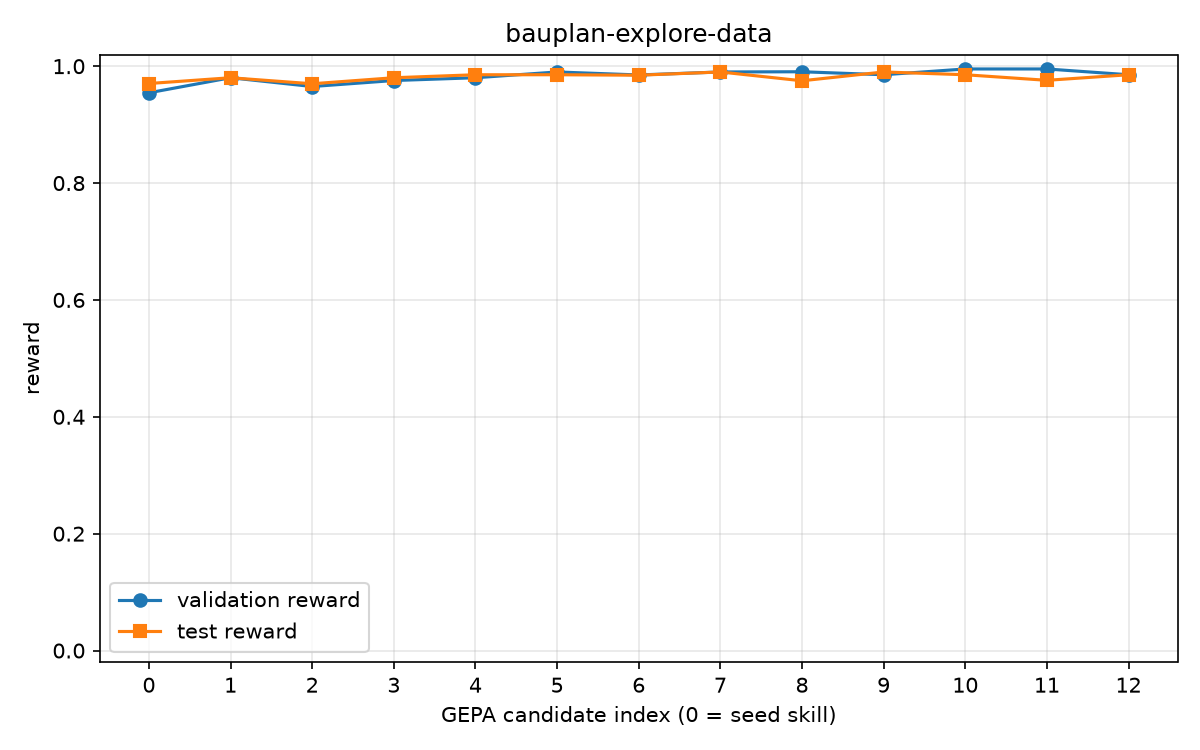}
    \caption{bauplan-explore-data}
    \label{fig:skill4-explore-data}
  \end{subfigure}
    \hfill%
  \begin{subfigure}[t]{0.32\textwidth}
    \includegraphics[width=\linewidth]{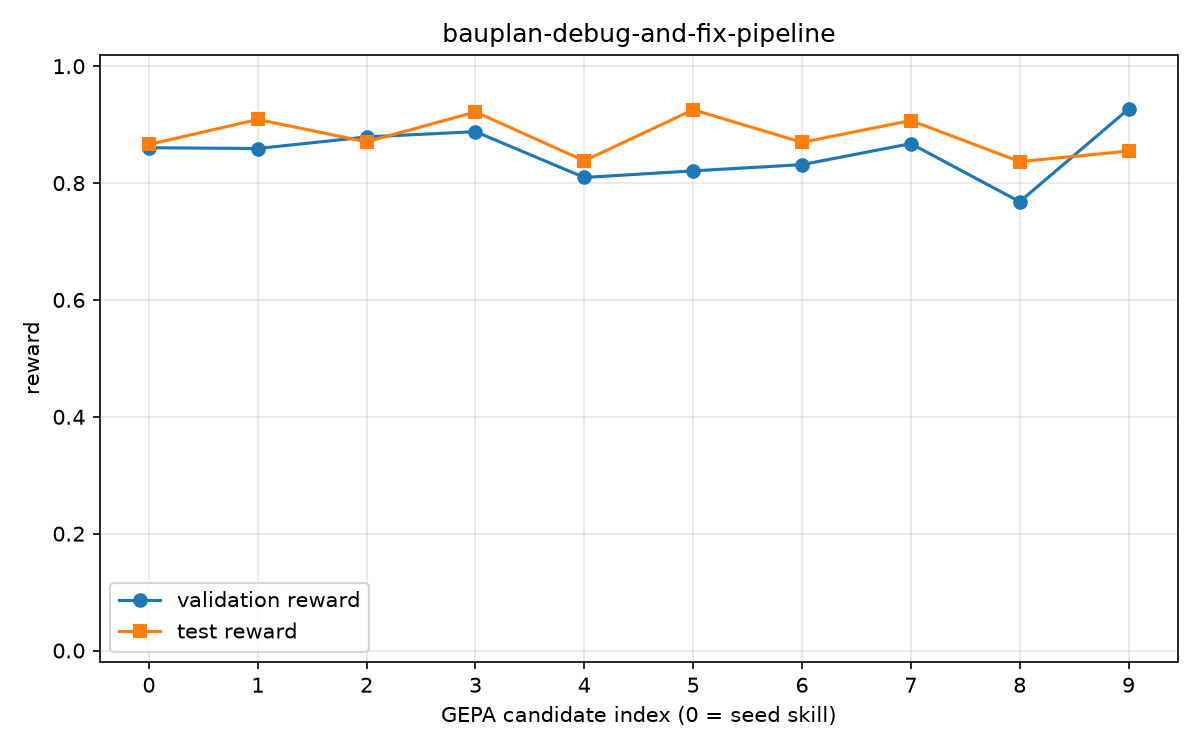}
    \caption{bauplan-debug-and-fix-pipeline}
    \label{fig:skill4-debug-and-fix}
  \end{subfigure}
  \hfill
  \begin{subfigure}[t]{0.32\textwidth}
    \includegraphics[width=\linewidth]{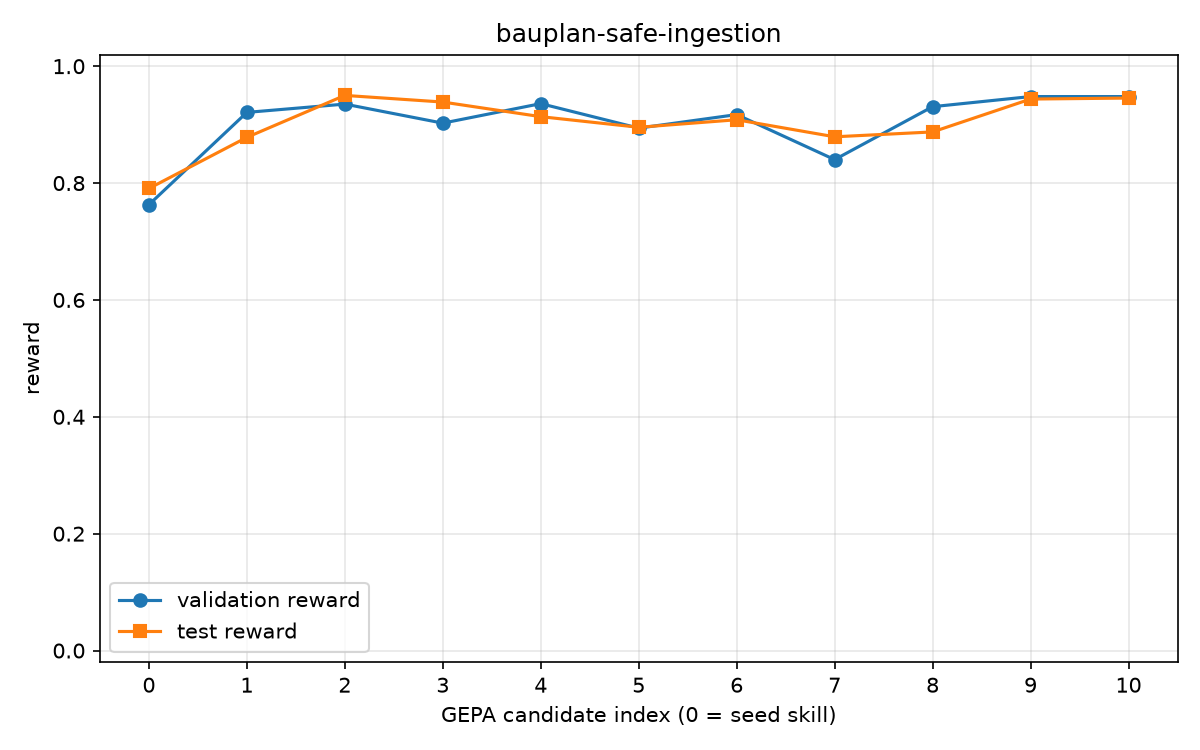}
    \caption{bauplan-safe-ingestion}
    \label{fig:skill4-safe-ingestion}
  \end{subfigure}

  \caption{GEPA optimization curves for six skills: validation and held-out test reward by candidate index.}
  \label{fig:skill-curves}
\end{figure*}

\section{Experiments}

Our final dataset is summarized in Table~\ref{tab:dataset}, comprising a total of 780 tasks with varying degrees of difficulty (Appendix~\ref{app:examples}). For each of the 6 skills we attempt to optimize, we extract the pool of dataset entries that require the given skill, then split the pool into train, validation, and test sets using a $70\%-15\%-15\%$ split. GEPA reflects and mutates on the train set and uses the validation set to select its best candidate. The test set is held out and used to evaluate performance. Because we evaluate only one coding-agent/LLM configuration, we consider these results important but preliminary evidence of the usefulness of the optimization loop.

Our primary metric is reward (defined as the \textit{fraction of validation checks passed}) measured on the validation and held-out test sets. We evaluate this metric for every candidate skill GEPA generates. The task LLM is \textit{claude-sonnet-4-6}, with Claude Code as the assistant. The initial set of skills comprises six different skills: \textit{bauplan-data-pipeline}, \textit{bauplan-safe-ingestion}, \textit{bauplan-data-quality-checks}, \textit{bauplan-data-assessment}, \textit{bauplan-explore-data}, \textit{bauplan-debug-and-fix-pipeline}.

Results are shown in Fig.~\ref{fig:skill-curves}. For every skill, GEPA found a candidate that not only performs better on the validation tasks, but also improves held-out reward (relative to the seed skills) by 12.0\% on average, suggesting that the optimized skills generalize beyond the validation tasks.

Qualitatively, the optimization process adds detail to the skill and appropriately edits the skill descriptions when validation checks fail due to the agent not invoking the expected skill. On average, the optimized skills were $27.2$ times longer (in terms of character count) than the seed skills.
Improvements included adding pointers to documentation, phrases to encourage the agent not to invent or guess flags, and general good practices (e.g., \texttt{Write-Audit-Publish on an isolated branch}).

In the cases where GEPA found only a small improvement in reward score, we identified that the dataset tasks associated with those skills were multi-skill tasks, where solving the task required correctly leveraging multiple skills. This result points to a clear line of future work: jointly optimizing skills. Our approach optimizes each skill in isolation, but the dataset supports complex tasks where skills may need to be adapted jointly to enable the agent to complete a correct workflow, end-to-end.


\section{Related work}
Recent work has optimized skills across tasks as diverse as software engineering, question answering, mathematical reasoning, web navigation, and tool use \cite{Gong2026skillmoo,Alzubi2026evoskill,Hwang2026agentpso,Vishe2026skill,Gao2026skillreducer,Huang2026bilevel}. Work on synthetic data generation and executable benchmarks similarly shows the value of generated tasks with verifiable outcomes \cite{wang2023selfinstruct,davidson2026reasoningdrivensyntheticdatageneration,jimenez2024swebench,yao2025taubench}. Our setting differs in being data-centric and lakehouse-specific: skills are optimized for concrete data engineering workflows such as ingestion, pipeline construction and debugging. Unlike read-path evaluations such as data-science agents \cite{liu2025surveytexttosqlerallms,chen2025largelanguagemodelbaseddata}, our benchmark is grounded in end-to-end data management use cases and traces from an agent-first production system at scale. In particular, we make the key observation that fine-grained evaluation of potentially disruptive data operations is possible only thanks to the isomorphism between agent-generated pipeline code and observable lakehouse state.
\section{Conclusion and future work}

We introduce a data-centric methodology for skill optimization serving data agents on a branching lakehouse. Combining real-world production traces, purpose-built agentic infrastructure, domain knowledge and a black-box optimizer, we provide preliminary evidence that this process can automatically improve compound AI systems. To the best of our knowledge, we have shown for the first time how to perform verification at scale for destructive data engineering tasks on production data: reliable, programmatic scoring of complex tasks would not have  been possible without Bauplan's vertical integration of agentic affordances (everything-as-code and git-for-data, Section~\ref{sec:bauplan}). By sharing our data generation process and releasing an open-source implementation, we believe our methodology can easily generalize to a broader set of use cases. For example, in ongoing work within our lab we are leveraging this process to fine-tune coding models: as LLMs are prone to reward hacking~\cite{wen2024languagemodelslearnmislead}, fine-grained verification is a precondition for effective reinforcement learning.

\bibliographystyle{ACM-Reference-Format}
\bibliography{sample}

@inproceedings{10.1145/3650203.3663335,
author = {Tagliabue, Jacopo and Greco, Ciro},
title = {Reproducible data science over data lakes: replayable data pipelines with Bauplan and Nessie},
year = {2024},
isbn = {9798400706110},
publisher = {Association for Computing Machinery},
address = {New York, NY, USA},
url = {https://doi.org/10.1145/3650203.3663335},
doi = {10.1145/3650203.3663335},
booktitle = {Proceedings of the Eighth Workshop on Data Management for End-to-End Machine Learning},
pages = {67–71},
location = {Santiago, AA, Chile},
series = {DEEM '24}
}

@misc{liu2025surveytexttosqlerallms,
      title={A Survey of Text-to-SQL in the Era of LLMs: Where are we, and where are we going?}, 
      author={Xinyu Liu and Shuyu Shen and Boyan Li and Peixian Ma and Runzhi Jiang and Yuxin Zhang and Ju Fan and Guoliang Li and Nan Tang and Yuyu Luo},
      year={2025},
      eprint={2408.05109},
      archivePrefix={arXiv},
      primaryClass={cs.DB},
      url={https://arxiv.org/abs/2408.05109}, 
}

@misc{chen2025largelanguagemodelbaseddata,
      title={Large Language Model-based Data Science Agent: A Survey}, 
      author={Ke Chen and Peiran Wang and Yaoning Yu and Xianyang Zhan and Haohan Wang},
      year={2025},
      eprint={2508.02744},
      archivePrefix={arXiv},
      primaryClass={cs.AI},
      url={https://arxiv.org/abs/2508.02744}, 
}

@misc{sheng2026buildingcorrectbydesignlakehousedata,
      title={Building a Correct-by-Design Lakehouse. Data Contracts, Versioning, and Transactional Pipelines for Humans and Agents}, 
      author={Weiming Sheng and Jinlang Wang and Manuel Barros and Aldrin Montana and Jacopo Tagliabue and Luca Bigon},
      year={2026},
      eprint={2602.02335},
      archivePrefix={arXiv},
      primaryClass={cs.DC},
      url={https://arxiv.org/abs/2602.02335}, 
}

@inproceedings{10.1145/3702634.3702955,
author = {Tagliabue, Jacopo and Caraza-Harter, Tyler and Greco, Ciro},
title = {Bauplan: Zero-copy, Scale-up FaaS for Data Pipelines},
year = {2024},
isbn = {9798400713361},
publisher = {Association for Computing Machinery},
address = {New York, NY, USA},
url = {https://doi.org/10.1145/3702634.3702955},
doi = {10.1145/3702634.3702955},
booktitle = {Proceedings of the 10th International Workshop on Serverless Computing},
pages = {31–36},
numpages = {6},
location = {Hong Kong, Hong Kong},
series = {WoSC10 '24}
}

@misc{iceberg,
  author = {Apache},
  title = {Iceberg},
  year = {2024},
  publisher = {GitHub},
  journal = {GitHub repository},
  howpublished = {\url{https://github.com/apache/iceberg}},
}

@misc{davidson2026reasoningdrivensyntheticdatageneration,
      title={Reasoning-Driven Synthetic Data Generation and Evaluation}, 
      author={Tim R. Davidson and Benoit Seguin and Enrico Bacis and Cesar Ilharco and Hamza Harkous},
      year={2026},
      eprint={2603.29791},
      archivePrefix={arXiv},
      primaryClass={cs.AI},
      url={https://arxiv.org/abs/2603.29791}, 
}

@misc{sheng2026gitlakegitfordataagenticlakehouse,
      title={GitLake: Git-for-data for the agentic lakehouse}, 
      author={Weiming Sheng and Jinlang Wang and Manuel Barros and Aldrin Montana and Jacopo Tagliabue and Luca Bigon},
      year={2026},
      eprint={2607.08319},
      archivePrefix={arXiv},
      primaryClass={cs.DB},
      url={https://arxiv.org/abs/2607.08319}, 
}

@INPROCEEDINGS {10825377,
author = { Tagliabue, Jacopo and Curtin, Ryan and Greco, Ciro },
booktitle = { 2024 IEEE International Conference on Big Data (BigData) },
title = {{ FaaS and Furious: abstractions and differential caching for efficient data pre-processing}},
year = {2024},
volume = {},
ISSN = {},
pages = {3562-3567},
doi = {10.1109/BigData62323.2024.10825377},
url = {https://doi.ieeecomputersociety.org/10.1109/BigData62323.2024.10825377},
publisher = {IEEE Computer Society},
address = {Los Alamitos, CA, USA},
month =Dec}

@misc{ant_skill,
  author = {Anthropic},
  title = {The Complete Guide to Building Skills for Claude},
  url = {https://resources.anthropic.com/hubfs/The-Complete-Guide-to-Building-Skill-for-Claude.pdf},
  year= {2025}
}

@misc{chen2025optimizingmodelselectioncompound,
      title={Optimizing Model Selection for Compound AI Systems}, 
      author={Lingjiao Chen and Jared Quincy Davis and Boris Hanin and Peter Bailis and Matei Zaharia and James Zou and Ion Stoica},
      year={2025},
      eprint={2502.14815},
      archivePrefix={arXiv},
      primaryClass={cs.AI},
      url={https://arxiv.org/abs/2502.14815}, 
}

@inproceedings{wang2023selfinstruct,
  title = {Self-Instruct: Aligning Language Models with Self-Generated Instructions},
  author = {Wang, Yizhong and Kordi, Yeganeh and Mishra, Swaroop and Liu, Alisa and Smith, Noah A. and Khashabi, Daniel and Hajishirzi, Hannaneh},
  booktitle = {Proceedings of the 61st Annual Meeting of the Association for Computational Linguistics},
  year = {2023},
  doi = {10.18653/v1/2023.acl-long.754}
}

@misc{montana2026notyourusualtypes,
  title        = {Not Your Usual Type(s): Data Contracts as Types Across Languages and Engines},
  author       = {Montana, Aldrin and Marc, Colin and Bigon, Luca and Tagliabue, Jacopo},
  year         = {2026},
  howpublished = {Preprint}
}

@inproceedings{jimenez2024swebench,
  title = {{SWE}-bench: Can Language Models Resolve Real-World {GitHub} Issues?},
  author = {Jimenez, Carlos E. and Yang, John and Wettig, Alexander and Yao, Shunyu and Pei, Kexin and Press, Ofir and Narasimhan, Karthik},
  booktitle = {The Twelfth International Conference on Learning Representations},
  year = {2024}
}

@inproceedings{zheng2023judging,
  title = {Judging {LLM}-as-a-Judge with {MT}-Bench and Chatbot Arena},
  author = {Zheng, Lianmin and Chiang, Wei-Lin and Sheng, Ying and Zhuang, Siyuan and Wu, Zhanghao and Zhuang, Yonghao and Lin, Zi and Li, Zhuohan and Li, Dacheng and Xing, Eric P. and Zhang, Hao and Gonzalez, Joseph E. and Stoica, Ion},
  booktitle = {Advances in Neural Information Processing Systems},
  year = {2023}
}

@inproceedings{yao2025taubench,
  title = {$\tau$-bench: A Benchmark for Tool-Agent-User Interaction in Real-World Domains},
  author = {Yao, Shunyu and Shinn, Noah and Razavi, Pedram and Narasimhan, Karthik},
  booktitle = {The Thirteenth International Conference on Learning Representations},
  year = {2025}
}

@misc{zha2023data,
      title={Data-centric Artificial Intelligence: A Survey}, 
      author={Daochen Zha and Zaid Pervaiz Bhat and Kwei-Herng Lai and Fan Yang and Zhimeng Jiang and Shaochen Zhong and Xia Hu},
      year={2023},
      eprint={2303.10158},
      archivePrefix={arXiv},
      primaryClass={cs.LG},
      url={https://arxiv.org/abs/2303.10158}, 
}

@misc{chen2025theorymindlargelanguage,
      title={Theory of Mind in Large Language Models: Assessment and Enhancement}, 
      author={Ruirui Chen and Weifeng Jiang and Chengwei Qin and Cheston Tan},
      year={2025},
      eprint={2505.00026},
      archivePrefix={arXiv},
      primaryClass={cs.CL},
      url={https://arxiv.org/abs/2505.00026}, 
}

@misc{liu2026diveclaudecodedesign,
      title={Dive into Claude Code: The Design Space of Today's and Future AI Agent Systems}, 
      author={Jiacheng Liu and Xiaohan Zhao and Xinyi Shang and Zhiqiang Shen},
      year={2026},
      eprint={2604.14228},
      archivePrefix={arXiv},
      primaryClass={cs.SE},
      url={https://arxiv.org/abs/2604.14228}, 
}

@misc{liu2026evoxmetaevolutionautomateddiscovery,
      title={EvoX: Meta-Evolution for Automated Discovery}, 
      author={Shu Liu and Shubham Agarwal and Monishwaran Maheswaran and Mert Cemri and Zhifei Li and Qiuyang Mang and Ashwin Naren and Ethan Boneh and Audrey Cheng and Melissa Z. Pan and Alexander Du and Kurt Keutzer and Alvin Cheung and Alexandros G. Dimakis and Koushik Sen and Matei Zaharia and Ion Stoica},
      year={2026},
      eprint={2602.23413},
      archivePrefix={arXiv},
      primaryClass={cs.LG},
      url={https://arxiv.org/abs/2602.23413}, 
}

@misc{novikov2025alphaevolvecodingagentscientific,
      title={AlphaEvolve: A coding agent for scientific and algorithmic discovery}, 
      author={Alexander Novikov and Ngân Vũ and Marvin Eisenberger and Emilien Dupont and Po-Sen Huang and Adam Zsolt Wagner and Sergey Shirobokov and Borislav Kozlovskii and Francisco J. R. Ruiz and Abbas Mehrabian and M. Pawan Kumar and Abigail See and Swarat Chaudhuri and George Holland and Alex Davies and Sebastian Nowozin and Pushmeet Kohli and Matej Balog},
      year={2025},
      eprint={2506.13131},
      archivePrefix={arXiv},
      primaryClass={cs.AI},
      url={https://arxiv.org/abs/2506.13131}, 
}

@misc{wen2024languagemodelslearnmislead,
      title={Language Models Learn to Mislead Humans via RLHF}, 
      author={Jiaxin Wen and Ruiqi Zhong and Akbir Khan and Ethan Perez and Jacob Steinhardt and Minlie Huang and Samuel R. Bowman and He He and Shi Feng},
      year={2024},
      eprint={2409.12822},
      archivePrefix={arXiv},
      primaryClass={cs.CL},
      url={https://arxiv.org/abs/2409.12822}, 
}

@misc{gelp2025machinetheorymindlarge,
      title={Towards Machine Theory of Mind with Large Language Model-Augmented Inverse Planning}, 
      author={Rebekah A. Gelpí and Eric Xue and William A. Cunningham},
      year={2025},
      eprint={2507.03682},
      archivePrefix={arXiv},
      primaryClass={cs.AI},
      url={https://arxiv.org/abs/2507.03682}, 
}

@software{Harbor_Framework,
author = {{Harbor Framework Team}},
month = jan,
title = {{Harbor: A framework for evaluating and optimizing agents and models in container environments}},
url = {https://github.com/harbor-framework/harbor},
year = {2026}
}

@article{baker2009action,
  title={Action understanding as inverse planning},
  author={Baker, Chris L and Saxe, Rebecca and Tenenbaum, Joshua B},
  journal={Cognition},
  volume={113},
  number={3},
  pages={329--349},
  year={2009},
  publisher={Elsevier}
}

@misc{liang2026skillnetcreateevaluateconnect,
      title={SkillNet: Create, Evaluate, and Connect AI Skills}, 
      author={Yuan Liang and Ruobin Zhong and Haoming Xu and Chen Jiang and Yi Zhong and Runnan Fang and Jia-Chen Gu and Shumin Deng and Yunzhi Yao and Mengru Wang and Shuofei Qiao and Xin Xu and Tongtong Wu and Kun Wang and Yang Liu and Zhen Bi and Jungang Lou and Yuchen Eleanor Jiang and Hangcheng Zhu and Gang Yu and Haiwen Hong and Longtao Huang and Hui Xue and Chenxi Wang and Yijun Wang and Zifei Shan and Xi Chen and Zhaopeng Tu and Feiyu Xiong and Xin Xie and Peng Zhang and Zhengke Gui and Lei Liang and Jun Zhou and Chiyu Wu and Jin Shang and Yu Gong and Junyu Lin and Changliang Xu and Hongjie Deng and Wen Zhang and Keyan Ding and Qiang Zhang and Fei Huang and Ningyu Zhang and Jeff Z. Pan and Guilin Qi and Haofen Wang and Huajun Chen},
      year={2026},
      eprint={2603.04448},
      archivePrefix={arXiv},
      primaryClass={cs.AI},
      url={https://arxiv.org/abs/2603.04448}, 
}

@misc{xu2026agentskillslargelanguage,
      title={Agent Skills for Large Language Models: Architecture, Acquisition, Security, and the Path Forward}, 
      author={Renjun Xu and Yang Yan},
      year={2026},
      eprint={2602.12430},
      archivePrefix={arXiv},
      primaryClass={cs.MA},
      url={https://arxiv.org/abs/2602.12430}, 
}

@misc{liu2025supportingaioverlordsredesigning,
      title={Supporting Our AI Overlords: Redesigning Data Systems to be Agent-First}, 
      author={Shu Liu and Soujanya Ponnapalli and Shreya Shankar and Sepanta Zeighami and Alan Zhu and Shubham Agarwal and Ruiqi Chen and Samion Suwito and Shuo Yuan and Ion Stoica and Matei Zaharia and Alvin Cheung and Natacha Crooks and Joseph E. Gonzalez and Aditya G. Parameswaran},
      year={2025},
      eprint={2509.00997},
      archivePrefix={arXiv},
      primaryClass={cs.AI},
      url={https://arxiv.org/abs/2509.00997}, 
}

@misc{yao2023reactsynergizingreasoningacting,
      title={ReAct: Synergizing Reasoning and Acting in Language Models}, 
      author={Shunyu Yao and Jeffrey Zhao and Dian Yu and Nan Du and Izhak Shafran and Karthik Narasimhan and Yuan Cao},
      year={2023},
      eprint={2210.03629},
      archivePrefix={arXiv},
      primaryClass={cs.CL},
      url={https://arxiv.org/abs/2210.03629}, 
}

@misc{huang_control_2025,
  title        = {Professional Software Developers Don't Vibe, They Control: AI Agent Use for Coding in 2025},
  author       = {Huang, Ruanqianqian and Reyna, Avery and Lerner, Sorin and Xia, Haijun and Hempel, Brian},
  year         = {2025},
  eprint       = {2512.14012},
  archivePrefix= {arXiv},
  primaryClass = {cs.SE},
  url          = {https://arxiv.org/abs/2512.14012}
}

@inproceedings{
Agrawal2025gepa,
title={{GEPA}: Reflective Prompt Evolution Can Outperform Reinforcement Learning},
author={Lakshya A Agrawal and Shangyin Tan and Dilara Soylu and Noah Ziems and Rishi Khare and Krista Opsahl-Ong and Arnav Singhvi and Herumb Shandilya and Michael J Ryan and Meng Jiang and Christopher Potts and Koushik Sen and Alex Dimakis and Ion Stoica and Dan Klein and Matei Zaharia and Omar Khattab},
booktitle={First Workshop on Foundations of Reasoning in Language Models},
year={2025},
url={https://openreview.net/forum?id=4oo6XTL6Oj}
}

@article{Gong2026skillmoo,
  title={SkillMOO: Multi-Objective Optimization of Agent Skills for Software Engineering},
  author={Gong, Jingzhi and Gu, Ruizhen and Fei, Zhiwei and Cao, Yazhuo and Twist, Lukas and Geiger, Alina and Han, Shuo and Sobania, Dominik and Sarro, Federica and Zhang, Jie M},
  journal={arXiv preprint arXiv:2604.09297},
  year={2026}
}

@article{Alzubi2026evoskill,
  title={Evoskill: Automated skill discovery for multi-agent systems},
  author={Alzubi, Salaheddin and Provenzano, Noah and Bingham, Jaydon and Chen, Weiyuan and Vu, Tu},
  journal={arXiv preprint arXiv:2603.02766},
  year={2026}
}

@article{Hwang2026agentpso,
  title={AgentPSO: Evolving Agent Reasoning Skill via Multi-agent Particle Swarm Optimization},
  author={Hwang, Hyunmin and Kim, Jaemin and Kim, Choonghan and Chang, Hangeol and Ye, Jong Chul},
  journal={arXiv preprint arXiv:2605.08704},
  year={2026}
}

@article{Vishe2026skill,
  title={Skill-R1: Agent Skill Evolution via Reinforcement Learning},
  author={Vishe, Yash and Surana, Rohan and Jiang, Xunyi and Huang, Zihan and Li, Xintong and Kuang, Nikki Lijing and Yu, Tong and Rossi, Ryan A and Shang, Jingbo and McAuley, Julian and others},
  journal={arXiv preprint arXiv:2605.09359},
  year={2026}
}

@article{Gao2026skillreducer,
  title={Skillreducer: Optimizing llm agent skills for token efficiency},
  author={Gao, Yudong and Li, Zongjie and Ji, Zimo and Ma, Pingchuan and Wang, Shuai and others},
  journal={arXiv preprint arXiv:2603.29919},
  year={2026}
}

@article{Huang2026bilevel,
  title={Bilevel Optimization of Agent Skills via Monte Carlo Tree Search},
  author={Huang, Chenyi and Zhang, Haoting and Xu, Jingxu and Zheng, Zeyu and Lin, Yunduan},
  journal={arXiv preprint arXiv:2604.15709},
  year={2026}
}

\appendix

\section{Examples of tasks}
\label{app:examples}
We reproduce here some task entries from the final dataset, slightly edited for length and content.

\begin{jsonblock}{Read task --- \texttt{assess\_customer\_support\_ticket\_data} (\texttt{read\_explore})}
  {
    "task_name": "assess_customer_support_ticket_data",
    "category": "read_explore",
    "difficulty": "medium",
    "goal": "Evaluate the distribution of ticket categories in the customer support tickets dataset and identify any anomalies in the data.",
    "prompt": "Can you help me analyze the customer support tickets to understand the distribution of ticket categories and check for any data anomalies?",
    "initial_state": {
      "branches": [
        "main"
      ],
      "seeded_tables": [
        {
          "table": "support.tickets",
          "branch": "main"
        }
      ]
    },
    "expected_commands": [
      "bauplan query"
    ],
    "expected_skills": [
      "bauplan-explore-data"
    ],
    "end_conditions": {
      "hard_response_constraints": [
        {
          "kind": "contains",
          "value": "distribution of ticket categories",
          "description": "Response should include analysis of ticket categories."
        }
      ],
      "soft_response_constraints": [
        {
          "criterion": "category_distribution_is_quantitative",
          "description": "The reply should describe the category distribution in concrete terms - counts or proportions per category, or at minimum a ranking of categories from most to least common - not a vague 'there are several categories' summary. A judge should see evidence the model actually grouped the data."
        }
      ],
      "state_assertions": [
        {
          "kind": "no_change",
          "target": "support.tickets",
          "expected": null,
          "description": "The ticket data should remain unchanged after exploration."
        }
      ]
    },
    "forbidden_actions": [
      {
        "pattern": "bauplan\\s+run\\b",
        "reason": "No new pipeline should be created during this task."
      }
    ]
}
\end{jsonblock}

Because this is a \textit{read}-only task, note that the constraints on the response are qualitative checks that leverage an LLM-as-a-judge at runtime.

On the other hand, a \textit{write} task such as the following includes detailed programmatic verification through a Python script (parts of which we report for convenience in a separate listing at the end): the script exits with status 0 on success or with a nonzero status (and an English explanation) on failure.

\begin{jsonblock}{Write task --- \texttt{build\_subway\_od\_pair\_summary\_pipeline} (\texttt{write\_new\_pipeline})}
  {
    "task_name": "build_subway_od_pair_summary_pipeline",
    "difficulty": "medium",
    "prompt_specificity": "detailed",
    "goal": "Build and publish a station origin destination aggregate pipeline from the existing MTA subway trips table onto the agent's personal main
  branch using a feature-branch workflow, without modifying global main.",
    "prompt": "Create a Bauplan project that reads the existing `transit.mta_subway_trips` table and materializes a new table named
  `analytics.subway_od_pair_summary` on your personal main lineage. Use the required branch workflow: discover your username programmatically, create a
  feature branch off your `<username>.main` branch, build and validate the output on that feature branch, then merge the successful feature branch back
  into your `<username>.main` branch. Do not write to the global `main` branch.\n\nThe output table `analytics.subway_od_pair_summary` should have one row
  per origin/destination station-complex pair after excluding rows where `origin_station_complex_id = destination_station_complex_id` and excluding rows
  with null or non-positive `estimated_average_ridership`. Include these columns:\n- `origin_station_complex_id`\n- `origin_station_complex_name`\n-
  `destination_station_complex_id`\n- `destination_station_complex_name`\n- `trip_record_count` = count of contributing source records\n-
  `total_estimated_ridership` = sum of `estimated_average_ridership`\n- `avg_estimated_ridership` = average of `estimated_average_ridership`\n\nBefore
  merging, validate that the table is readable, non-empty, has no same-station origin/destination pairs, and has positive total ridership. After merging,
  reply briefly with the destination table name and confirm it is on `<username>.main`.",
    "expected_commands": [
      "bauplan branch create",
      "bauplan run",
      "bauplan branch merge"
    ],
    "expected_skills": [
      "bauplan-data-pipeline",
      "bauplan-data-quality-checks"
    ],
    "end_conditions": {
      "hard_response_constraints": [
        {
          "kind": "contains",
          "value": "analytics.subway_od_pair_summary",
          "description": "Final reply names the materialized destination table."
        }
      ],
      "state_assertions": [
        {
          "kind": "no_change",
          "target": "main",
          "expected": null,
          "description": "The global main branch is unchanged by the write workflow."
        }
      ]
    },
    "validation_script": "<see Listing below>",
    "category": "write_new_pipeline"
  }
  \end{jsonblock}

\begin{lstlisting}[
  language=Python,
  showstringspaces=false,
  columns=fullflexible,
  caption={Sample validation script using the SDK.},
  label={lst:verification},
  basicstyle=\ttfamily\scriptsize,
  numbers=none
]
import sys

import bauplan


DEST_NAMESPACE = "analytics"
DEST_TABLE = "subway_od_pair_summary"
DEST_FULL_NAME = f"{DEST_NAMESPACE}.{DEST_TABLE}"


def fail(message: str) -> None:
    print(message, file=sys.stderr)
    raise SystemExit(1)


# Discover the agent's personal main branch.
try:
    client = bauplan.Client()
    info = client.info()
except Exception as exc:
    fail(f"Could not initialize the Bauplan client: {exc}")

if info.user is None or not info.user.username:
    fail("Could not discover the current Bauplan username")

username = info.user.username
user_main = f"{username}.main"


# The destination table must exist on the agent's personal main branch.
try:
    exists_on_user_main = client.has_table(
        DEST_TABLE,
        namespace=DEST_NAMESPACE,
        ref=user_main,
    )
except Exception as exc:
    fail(f"Could not inspect {DEST_FULL_NAME} on {user_main}: {exc}")

if not exists_on_user_main:
    fail(f"{DEST_FULL_NAME} does not exist on {user_main}")


# The destination table must not have been created on global main.
# The benchmark harness separately verifies that global main's commit
# did not change during the task.
try:
    exists_on_global_main = client.has_table(
        DEST_TABLE,
        namespace=DEST_NAMESPACE,
        ref="main",
    )
except Exception as exc:
    fail(f"Could not inspect {DEST_FULL_NAME} on global main: {exc}")

if exists_on_global_main:
    fail(f"{DEST_FULL_NAME} was created on global main")


# Check that every required output column is present.
required_columns = {
    "origin_station_complex_id",
    "origin_station_complex_name",
    "destination_station_complex_id",
    "destination_station_complex_name",
    "trip_record_count",
    "total_estimated_ridership",
    "avg_estimated_ridership",
}

try:
    schema_table = client.query(
        f"SELECT * FROM {DEST_FULL_NAME} LIMIT 0",
        ref=user_main,
    )
except Exception as exc:
    fail(f"Could not inspect the schema of {DEST_FULL_NAME}: {exc}")

actual_columns = {name.lower() for name in schema_table.column_names}
missing_columns = sorted(required_columns - actual_columns)

if missing_columns:
    fail(
        "Destination table is missing required columns: "
        + ", ".join(missing_columns)
    )

# Recompute the expected result from the source table and compare the
# complete expected and actual relations. Numeric aggregates are rounded
# only for comparison, to tolerate insignificant floating-point differences.
validation_query = f"""
WITH expected AS (
    SELECT
        origin_station_complex_id,
        origin_station_complex_name,
        destination_station_complex_id,
        destination_station_complex_name,
        COUNT(*) AS trip_record_count,
        SUM(estimated_average_ridership)
            AS total_estimated_ridership,
        AVG(estimated_average_ridership)
            AS avg_estimated_ridership
    FROM transit.mta_subway_trips
    WHERE estimated_average_ridership IS NOT NULL
      AND estimated_average_ridership > 0
      AND origin_station_complex_id
          <> destination_station_complex_id
    GROUP BY
        origin_station_complex_id,
        origin_station_complex_name,
        destination_station_complex_id,
        destination_station_complex_name
),
actual AS (
    SELECT
        origin_station_complex_id,
        origin_station_complex_name,
        destination_station_complex_id,
        destination_station_complex_name,
        trip_record_count,
        total_estimated_ridership,
        avg_estimated_ridership
    FROM {DEST_FULL_NAME}
),
expected_normalized AS (
    SELECT
        origin_station_complex_id,
        origin_station_complex_name,
        destination_station_complex_id,
        destination_station_complex_name,
        trip_record_count,
        ROUND(total_estimated_ridership, 6)
            AS total_estimated_ridership,
        ROUND(avg_estimated_ridership, 6)
            AS avg_estimated_ridership
    FROM expected
),
actual_normalized AS (
    SELECT
        origin_station_complex_id,
        origin_station_complex_name,
        destination_station_complex_id,
        destination_station_complex_name,
        trip_record_count,
        ROUND(total_estimated_ridership, 6)
            AS total_estimated_ridership,
        ROUND(avg_estimated_ridership, 6)
            AS avg_estimated_ridership
    FROM actual
),
missing_rows AS (
    SELECT *
    FROM expected_normalized

    EXCEPT ALL

    SELECT *
    FROM actual_normalized
),
unexpected_rows AS (
    SELECT *
    FROM actual_normalized

    EXCEPT ALL

    SELECT *
    FROM expected_normalized
)
SELECT
    (SELECT COUNT(*) FROM expected) AS expected_rows,
    (SELECT COUNT(*) FROM actual) AS actual_rows,
    (SELECT COUNT(*) FROM missing_rows) AS missing_rows,
    (SELECT COUNT(*) FROM unexpected_rows) AS unexpected_rows,

    (
        SELECT COUNT(*)
        FROM actual
        WHERE origin_station_complex_id
            = destination_station_complex_id
    ) AS same_station_rows,

    (
        SELECT COUNT(*)
        FROM actual
        WHERE trip_record_count IS NULL
           OR trip_record_count <= 0
    ) AS invalid_trip_count_rows,

    (
        SELECT COUNT(*)
        FROM actual
        WHERE total_estimated_ridership IS NULL
           OR total_estimated_ridership <= 0
    ) AS invalid_total_rows,

    (
        SELECT COUNT(*)
        FROM actual
        WHERE avg_estimated_ridership IS NULL
           OR avg_estimated_ridership <= 0
    ) AS invalid_average_rows
"""

try:
    result = client.query(
        validation_query,
        ref=user_main,
        max_rows=1,
    )
except Exception as exc:
    fail(f"Validation query failed: {exc}")

if result.num_rows != 1:
    fail("Validation query did not return exactly one result row")

checks = result.to_pylist()[0]


if checks["actual_rows"] <= 0:
    fail(f"{DEST_FULL_NAME} is empty")

if checks["same_station_rows"] != 0:
    fail(
        "Destination table contains "
        f"{checks['same_station_rows']} same-station "
        "origin/destination rows"
    )

if checks["invalid_trip_count_rows"] != 0:
    fail(
        "Destination table contains "
        f"{checks['invalid_trip_count_rows']} rows with a null or "
        "non-positive trip_record_count"
    )

if checks["invalid_total_rows"] != 0:
    fail(
        "Destination table contains "
        f"{checks['invalid_total_rows']} rows with null or "
        "non-positive total_estimated_ridership"
    )

if checks["invalid_average_rows"] != 0:
    fail(
        "Destination table contains "
        f"{checks['invalid_average_rows']} rows with null or "
        "non-positive avg_estimated_ridership"
    )

if checks["actual_rows"] != checks["expected_rows"]:
    fail(
        "Row-count mismatch: "
        f"expected {checks['expected_rows']}, "
        f"found {checks['actual_rows']}"
    )

if checks["missing_rows"] != 0:
    fail(
        "Destination table is missing "
        f"{checks['missing_rows']} expected rows"
    )

if checks["unexpected_rows"] != 0:
    fail(
        "Destination table contains "
        f"{checks['unexpected_rows']} incorrect or unexpected rows"
    )

print(
    f"PASS: {DEST_FULL_NAME} on {user_main} "
    "matches the expected origin/destination aggregation"
)
\end{lstlisting}


\end{document}